\pdfoutput=1
\documentclass[letterpaper, 10 pt, conference]{ieeeconf}  % Comment this line out if you need a4paper

\IEEEoverridecommandlockouts                              % This command is only needed if 
\usepackage{graphics} % for pdf, bitmapped graphics files
\usepackage{epsfig} % for postscript graphics files
\usepackage{amsmath} % assumes amsmath package installed
\usepackage{amssymb}  % assumes amsmath package installed
\usepackage{multirow} 
\usepackage{xcolor}

\usepackage{bm}
\usepackage{booktabs}
\usepackage{multirow}
\usepackage{makecell}
\usepackage{nicefrac}
\usepackage{graphicx}
\usepackage{colortbl}
\usepackage{algorithm}
\usepackage{algorithmic}

\usepackage{subcaption}
\usepackage{comment}
\usepackage{thmtools,thm-restate}
\usepackage{wrapfig}

\usepackage{amssymb} 

\makeatletter
\let\NAT@parse\undefined
\makeatother
\usepackage[breaklinks,colorlinks,citecolor=green]{hyperref}
\usepackage[capitalise]{cleveref}
\usepackage{cite}

\title{\LARGE \bf
SymDrive: Realistic and Controllable Driving Simulator via Symmetric Auto-regressive Online Restoration
}

\author{Zhiyuan Liu$^{1*}$, Daocheng Fu$^{2*}$, Pinlong Cai$^{2}$, Lening Wang$^{1,3}$, Ying Liu$^{1,\dagger}$,\\  Yilong Ren$^{3}$, Botian Shi$^{2}$, Jianqiang Wang$^{1}$%
\thanks{$^{1}$ School of Vehicle and Mobility, Tsinghua University.}%
\thanks{$^{2}$ Shanghai Artificial Intelligence Laboratory.}%
\thanks{$^{3}$ State Key Lab of Intelligent Transportation System, Beihang University}%
\thanks{$^\dagger$ Corresponding author: \tt\footnotesize seuliuy@hotmail.com} % <-this % stops a space
\thanks{$^{*}$ Equal contribution}
}

\begin{document}

\maketitle
\thispagestyle{empty}
\pagestyle{empty}

\begin{abstract}

High-fidelity and controllable 3D simulation is essential for addressing the long-tail data scarcity in Autonomous Driving (AD), yet existing methods struggle to simultaneously achieve photorealistic rendering and interactive traffic editing. Current approaches often falter in large-angle novel view synthesis and suffer from geometric or lighting artifacts during asset manipulation. To address these challenges, we propose SymDrive, a unified diffusion-based framework capable of joint high-quality rendering and scene editing. We introduce a Symmetric Auto-regressive Online Restoration paradigm, which constructs paired symmetric views to recover fine-grained details via a ground-truth-guided dual-view formulation and utilizes an auto-regressive strategy for consistent lateral view generation. Furthermore, we leverage this restoration capability to enable a training-free harmonization mechanism, treating vehicle insertion as context-aware inpainting to ensure seamless lighting and shadow consistency. Extensive experiments demonstrate that SymDrive achieves state-of-the-art performance in both novel-view enhancement and realistic 3D vehicle insertion.

\end{abstract}
\vspace{-2.0mm}

\section{Introduction}

% Autonomous driving (AD) technology has witnessed significant advancements and initial deployments in commercial vehicles. However, achieving Level 4 autonomy and beyond—requiring robust performance in complex, diverse scenarios—remains a formidable challenge. A primary bottleneck stems from the inherent limitations of data-driven AD models, particularly their susceptibility to the long-tail problem arising from the sparsity of critical events in real-world data collection\cite{Dauner2024NEURIPS, yang2024drivearena}. Consequently, generating high-fidelity and controllable 3D simulation environments has become an indispensable tool for comprehensively training and evaluating AD systems. For such simulations to be effective, two core capabilities are essential: \textbf{high-fidelity visual rendering} and \textbf{interactive scene editing}. Visual rendering must produce photorealistic, spatio-temporally coherent image sequences that satisfy the perception needs of the AD stack. Scene editing, on the other hand, must provide fine-grained control over traffic participants, including modifying their trajectories and manipulating traffic flow by adding or removing agents, while preserving photorealistic appearance and spatio-temporal consistency during editing, thereby enabling the creation of realistic and challenging traffic scenarios.

Despite the rapid commercial deployment of Autonomous Driving (AD) technology, achieving robust Level 4 autonomy remains impeded by the "long-tail" problem inherent in data-driven approaches, where critical edge cases are sparse in real-world datasets~\cite{Dauner2024NEURIPS, yang2024drivearena}. Consequently, high-fidelity, controllable 3D simulation has emerged as an imperative paradigm for comprehensively training and evaluating AD systems. To be effective, such simulators must satisfy two core requirements: \textbf{high-fidelity visual rendering} and \textbf{interactive scene editing}. The former demands the generation of photorealistic, spatio-temporally coherent image sequences tailored for perception models. The latter necessitates fine-grained control over traffic dynamics—such as modifying trajectories or adjusting agent density—while strictly maintaining visual and temporal consistency, thereby enabling the synthesis of diverse and challenging driving scenarios.

As summarized in~\cref{tab:simulator_comparison_color}, existing methods struggle to jointly satisfy visual fidelity and editable traffic. Video diffusion–based simulators~\cite{yang2024drivearena} offer realistic appearances but suffer from temporal inconsistency and slow inference. Conversely, 3D Gaussian Splatting (3DGS) approaches~\cite{kerbl3Dgaussians, chen2023periodic, yan2024street, chen2025omnire} achieve real-time rendering and strong consistency yet lack support for realistic traffic editing and generalization to novel views. While pixel editing models~\cite{podell2023sdxlimprovinglatentdiffusion, zhang2025scaling, ljungbergh2025r3d2realistic3dasset} allow for high-fidelity local modification, they cannot resolve the view-synthesis limitations of underlying 3D representations. Recent hybrid methods~\cite{wu2025difix3d+, yan2024streetcrafter, Ni2024ReconDreamerCW} combine 3DGS with diffusion priors to improve rendering; however, they have not explicitly addressed realistic traffic editing, and their novel-view synthesis quality remains suboptimal.

\newcommand{\cmark}{\textcolor{green!70!black}{\checkmark}}
\newcommand{\xmark}{\textcolor{red}{$\times$}}            

% \begin{table}[t]
% 	\centering
%     \caption{Comparison of features across different simulators.}
%     \vspace{-5pt}
% 	\resizebox{0.49\textwidth}{!}{
% 	\begin{tabular}{l c c c c c}
%     	\toprule
%     	\multirow{2}{*}{\textbf{Simulator}} & \multicolumn{3}{c}{\textbf{Visual Rendering}} & \multicolumn{2}{c}{\textbf{Scenario Editing}} \\
%         \cmidrule(lr){2-4} \cmidrule(lr){5-6}
%     	& \textbf{3D} & \textbf{Realistic} & \makecell{\textbf{Spatiotemporal} \\ \textbf{consistency}} & \makecell{\textbf{Trajectory} \\ \textbf{Editing}} & \makecell{\textbf{Traffic Volume} \\ \textbf{Control}} \\
%     	\midrule
%     	SUMO         & \xmark   & \xmark   & \xmark   & \cmark   & \cmark   \\
%     	CARLA        & \cmark   & \xmark   & \xmark   & \cmark   & \cmark   \\
%         MagicDrive & \cmark & \cmark & \xmark & \cmark & \cmark   \\
%     	DriveArena   & \cmark   & \cmark   & \xmark   & \cmark   & \cmark   \\
%         DriveStudio  & \cmark   & \cmark   & \cmark   & \xmark   & \xmark   \\
%         ReconDreamer & \cmark  & \cmark  & \cmark & \cmark & \xmark   \\
%         ReconDreamer++ & \cmark  & \cmark  & \cmark & \cmark & \xmark   \\
%     	Ours (AutoSim) & \cmark   & \cmark   & \cmark   & \cmark   & \cmark   \\
%     	\bottomrule
%     \end{tabular}
% 	}
% 	\label{tab:simulator_comparison_color}
%     \vspace{-1em}
% \end{table}

\newcommand{\Good}{\textcolor{green!60!black}{\scalebox{1.3}{\checkmark}}}
\newcommand{\Bad}{\textcolor{red}{\scalebox{1.3}{$\times$}}}
\newcommand{\Partial}{\textcolor{orange}{\scalebox{1.2}{$\triangle$}}}

\begin{table}[t]
  \centering
  \caption{Comparison of Controllable Traffic Scene Simulation Methods.
  \Good: fully supported; 
  \Partial: partially supported or limited; 
  \Bad: not supported.}

  \resizebox{0.49\textwidth}{!}{
  \renewcommand{\arraystretch}{1.05}
  \begin{tabular}{l l c c c c c}
    \toprule
    \textbf{Method} & \textbf{Method} 
    & \multirow{2}{*}{\textbf{Consistency}} 
    & \textbf{Trajectory} 
    & \textbf{Editing} 
    & \textbf{Novel View} 
    & \textbf{Real-time} \\
    
    \textbf{Category} & \textbf{Name} 
    & 
    & \textbf{Fidelity} 
    & \textbf{Realism} 
    & \textbf{Realism} 
    & \textbf{Rendering} \\
    \midrule

    \multirowcell{2}[0pt][l]{\textbf{Video} \\ \textbf{Diffusion}}
    & \multirow{2}{*}{DriveArena\cite{yang2024drivearena}}
    & \multirow{2}{*}{\Bad}
    & \multirow{2}{*}{\Partial}
    & \multirow{2}{*}{\Good}
    & \multirow{2}{*}{\Good} 
    & \multirow{2}{*}{\Bad} \\
    \\
    \midrule

    \multirow{3}{*}{\textbf{3DGS}}
    & PVG\cite{chen2023periodic}
    & \multirow{3}{*}{\Good} 
    & \multirow{3}{*}{\Good} 
    & \multirow{3}{*}{\Bad} 
    & \multirow{3}{*}{\Bad} 
    & \multirow{3}{*}{\Good} \\
    & StreetGS\cite{yan2024street} \\
    & OmniRe\cite{chen2025omnire} \\
    \midrule

    \multirowcell{3}[0pt][l]{\textbf{Edit} \\ \textbf{Models}}
    & CosXL-Edit\cite{podell2023sdxlimprovinglatentdiffusion} 
    & \multirow{3}{*}{\Bad} 
    & \multirow{3}{*}{\Good} 
    & \multirow{3}{*}{\Good} 
    & \multirow{3}{*}{\Bad} 
    & \multirow{3}{*}{\Partial} \\
    & IC-Light\cite{zhang2025scaling} \\
    & R3D2\cite{ljungbergh2025r3d2realistic3dasset} \\
    \midrule

    \multirowcell{3}[0pt][l]{\textbf{3DGS +} \\ \textbf{Diffusion}}
    & Difix3D\cite{wu2025difix3d+}
    & \multirow{3}{*}{\Good} 
    & \multirow{3}{*}{\Good} 
    & \multirow{3}{*}{\Bad} 
    & \multirow{3}{*}{\Partial} 
    & \multirow{3}{*}{\Good} \\
    & StreetCrafter\cite{yan2024streetcrafter} \\
    & ReconDreamer\cite{Ni2024ReconDreamerCW} \\
    \midrule

    \textbf{Ours}
    & -- 
    & \Good 
    & \Good 
    & \Good 
    & \Good 
    & \Good \\
    \bottomrule
  \end{tabular}
  }
  \label{tab:simulator_comparison_color}
  \vspace{-1em}
\end{table}

\begin{figure}[t]
    \centering
    \includegraphics[width=0.48\textwidth]{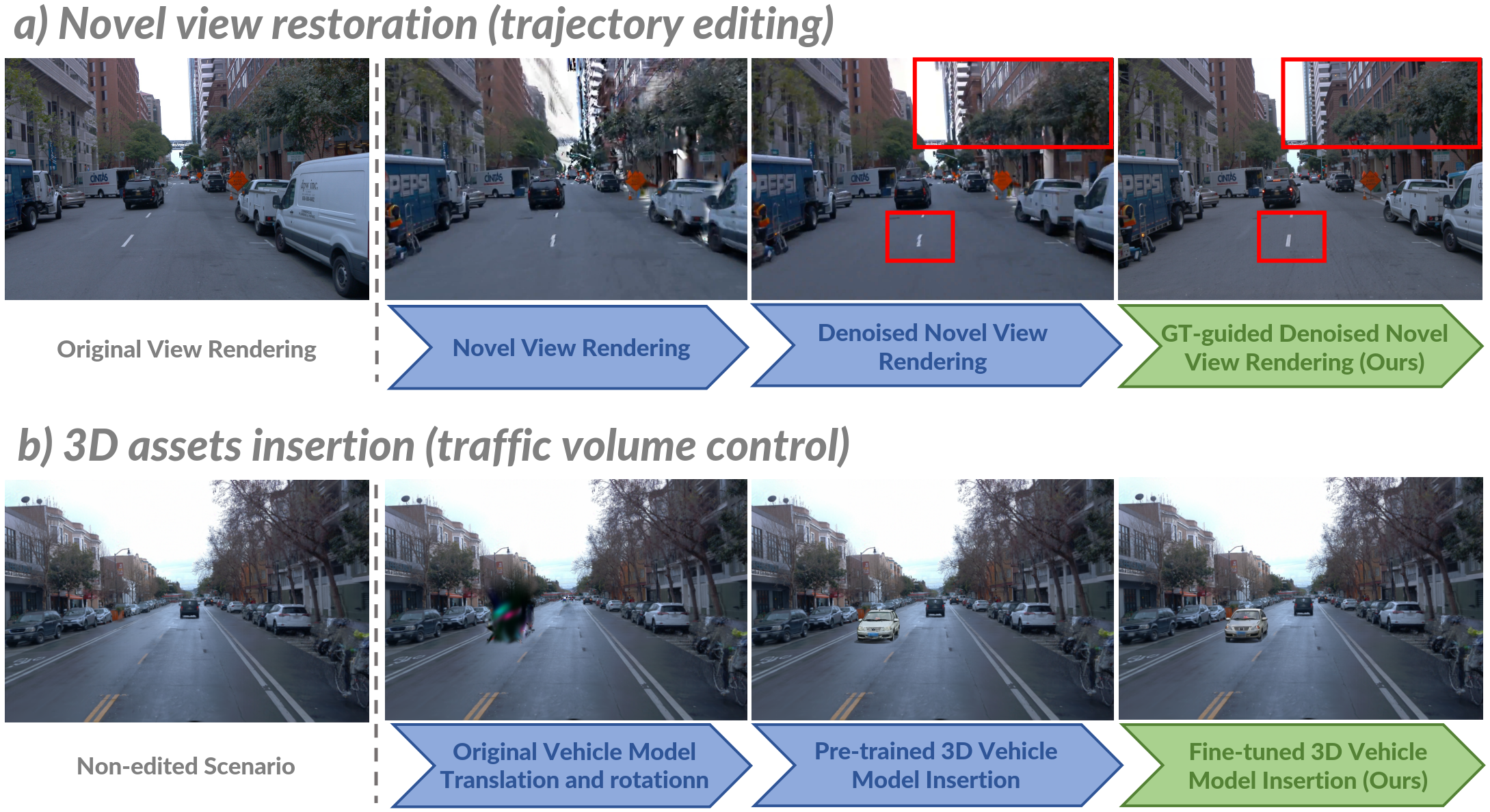}
    % \caption{Challenges of existing methods. a) Novel View Restoration: Current methods directly restore novel view renderings with missing information, leading to detail loss~\cite{fan2024freesimfreeviewpointcamerasimulation, Ni2024ReconDreamerCW, zhao2025recon}. Our autoregressive approach leverages Ground Truth (GT) image information, significantly enhancing detail fidelity. b) Model Integration in Scene Editing: Directly inserting pre-trained 3D models often causes color and illumination inconsistencies with the background~\cite{zhou2024hugsim}. We finetune the inserted models to reduce these discrepancies and improve fidelity.}
    \caption{Challenges of existing visual simulation for AD system. Enlarge the image to see details}
    \label{fig:problem_illustration}
    \vspace{-20pt}
\end{figure}

Two fundamental challenges constrain the deployment of current AD simulations. First, \textbf{high-quality novel-view synthesis remains unresolved} (see~\cref{fig:problem_illustration} a). Existing single-view restoration methods lack sufficient geometric constraints to recover details during lateral viewpoint shifts. Furthermore, reliance on costly tailored training data (e.g., masks~\cite{Ni2024ReconDreamerCW} or synthetic perturbations~\cite{fan2024freesimfreeviewpointcamerasimulation}) limits their scalability and effectiveness in real-world driving scenarios. Second, \textbf{realistic traffic editing faces severe artifacts} (see~\cref{fig:problem_illustration} b). Manipulating existing vehicles often exposes incomplete geometries, causing ghosting effects, while inserting new assets introduces lighting and shadowing inconsistencies, creating unnatural seams between foreground objects and the background.

To address these challenges, we propose a unified diffusion-based framework that jointly tackles both tasks. For novel-view synthesis, we depart from single-view inference by constructing paired symmetric views to recover the central ground truth (GT). This dual-view formulation leverages richer geometric and appearance priors to enhance restoration quality, while the symmetric design simplifies data generation. Furthermore, we implement lateral view synthesis via an auto-regressive strategy: starting from the GT, the model iteratively generates distant views by conditioning on the previous rendering. This effectively propagates scene details and preserves fine-grained consistency across viewpoints.

Leveraging the varied detail recovery capabilities of our model, we further introduce a training-free harmonization mechanism for traffic editing. We formulate vehicle insertion as a context-aware inpainting task: by masking the target region and conditioning on the surrounding context via dual-view inputs, the diffusion process naturally harmonizes the inserted vehicle's appearance, lighting, and shadows. This ensures seamless integration of synthesized assets, eliminating artifacts caused by geometric incompleteness or rendering inconsistencies.

Experimental results demonstrate that our method achieves state-of-the-art (SOTA) performance in both novel-view enhancement and 3D asset insertion. Our main contributions are summarized as follows:

\begin{itemize}
    \item We propose a unified framework that simultaneously handles novel-view synthesis and realistic traffic editing, eliminating the need for task-specific modules or separate training stages.
    \item We introduce a GT-guided online restoration paradigm featuring symmetric dual-view construction and auto-regressive lateral propagation, enabling accurate fine-grained detail recovery and efficient view generation.
    \item Extensive experiments demonstrate that our method achieves SOTA performance in novel-view enhancement and 3D vehicle insertion, validating its potential for realistic simulation environments.
\end{itemize}

% \begin{itemize}
%     \item \textbf{GT-Guided Denoising for Novel View Synthesis:} We propose a novel symmetric auto-regressive denoising technique that leverages ground truth (GT) images as strong priors, significantly improving novel view rendering quality by correcting artifacts while maintaining scene fidelity.
    
%     \item \textbf{Diffusion-Enhanced Traffic Editing:} We propose a novel diffusion-augmented solution for high-quality traffic scene editing, where the diffusion model is applied to restore the pre-trained 3D vehicle assets. This addresses the geometric incompleteness of vehicle models in traditional reconstruction methods while achieving superior background consistency compared to existing 3D vehicle model insertion approaches.
    
%     \item \textbf{High-Fidelity Closed-Loop Simulation:} Experimental results demonstrate state-of-the-art novel view rendering and effective vehicle editing capabilities. Combined with traffic controllers, our system enables controllable simulation of diverse traffic scenarios with high visual fidelity.
% \end{itemize}

\section{Related Work}\label{sec:related}

% Closed-loop simulation frameworks for autonomous driving systems integrate two essential components: visual and vehicle behavior simulation. Visual simulation provides high-fidelity sensory data, enabling the perception system to accurately interpret the operational environment. Simultaneously, vehicle behavior simulation generates dynamic and realistic traffic contexts, which are vital for validating the decision-making and planning capabilities of the autonomous vehicle in producing appropriate reactions.

\paragraph{Visual Rendering and Generation} Visual simulation for autonomous driving primarily employs two methodologies: neural rendering (e.g., NeRF \cite{mildenhall2021nerf, yang2024emernerf, yang2023unisim, tonderski2024neurad} and 3DGS \cite{kerbl3Dgaussians, chen2023periodic, huang2024s3gaussian, zhou2024drivinggaussian, yang2023deformable3dgs, yan2024street, chen2025omnire}) for reconstructing existing scenes, and generative models (e.g., diffusion models \cite{ho2020denoising, gao2023magicdrive, wang2024drivedreamer}) for synthesizing novel content. Neural rendering techniques excel at creating realistic 3D representations from 2D images, offering high spatio-temporal consistency and, with methods like 3DGS, real-time rendering. However, their fidelity degrades for novel viewpoints not well-covered by input data, leading to artifacts~\cite{he2024neural, liao2025learning}. Conversely, generative models can produce diverse, photorealistic scenes, including scenarios absent from training data, which is crucial for varied simulations. Yet, they often struggle with temporal consistency and incur high computational costs, challenging real-time, high-resolution generation~\cite{guan2024world}. Hybrid approaches are emerging to combine these strengths, using generative models' learned priors to enhance reconstruction-based outputs~\cite{Ni2024ReconDreamerCW, zhao2025recon}. This can involve denoising, refining, or in-painting renderings, especially for challenging novel views, thereby improving the quality, robustness, and realism of synthesized scenes for autonomous driving validation.

% \paragraph{Behavior Controlling} Behavior controlling in autonomous driving simulation employs distinct strategies, each with specific trade-offs. Traditional traffic flow simulators, such as SUMO \cite{SUMO2018} and Vissim \cite{ptv-vissim}, excel at modeling macroscopic traffic dynamics by ensuring aggregate parameters like flow and density adhere to established traffic constraints, though they often simplify the nuanced behaviors of individual vehicles. For more detailed interactions, rule-based vehicle control models, integral to platforms like CARLA \cite{Dosovitskiy17}, Apollo \cite{ap}, offer high-fidelity trajectory generation. These are governed by explicit algorithms that enforce vehicle dynamics and traffic laws, resulting in precise and compliant actions, but their deterministic nature may not fully capture the diverse decision-making of human drivers. Conversely, emerging data-driven vehicle control models, exemplified by Controllable Traffic Generation (CTG) \cite{10161463, zhong2023languageguided} and techniques like Dense Deep Reinforcement Learning (D2RL) \cite{feng2023dense}, learn complex behaviors from large-scale human driving datasets. This approach can produce more naturalistic, diverse, and contextually appropriate agent actions, significantly enhancing simulation realism. However, the ``black box'' nature of these learned models often introduces challenges in interpretability and formal safety verification, particularly for out-of-distribution scenarios.

\begin{figure*}[t]
    \centering
    \includegraphics[width=0.97\textwidth]{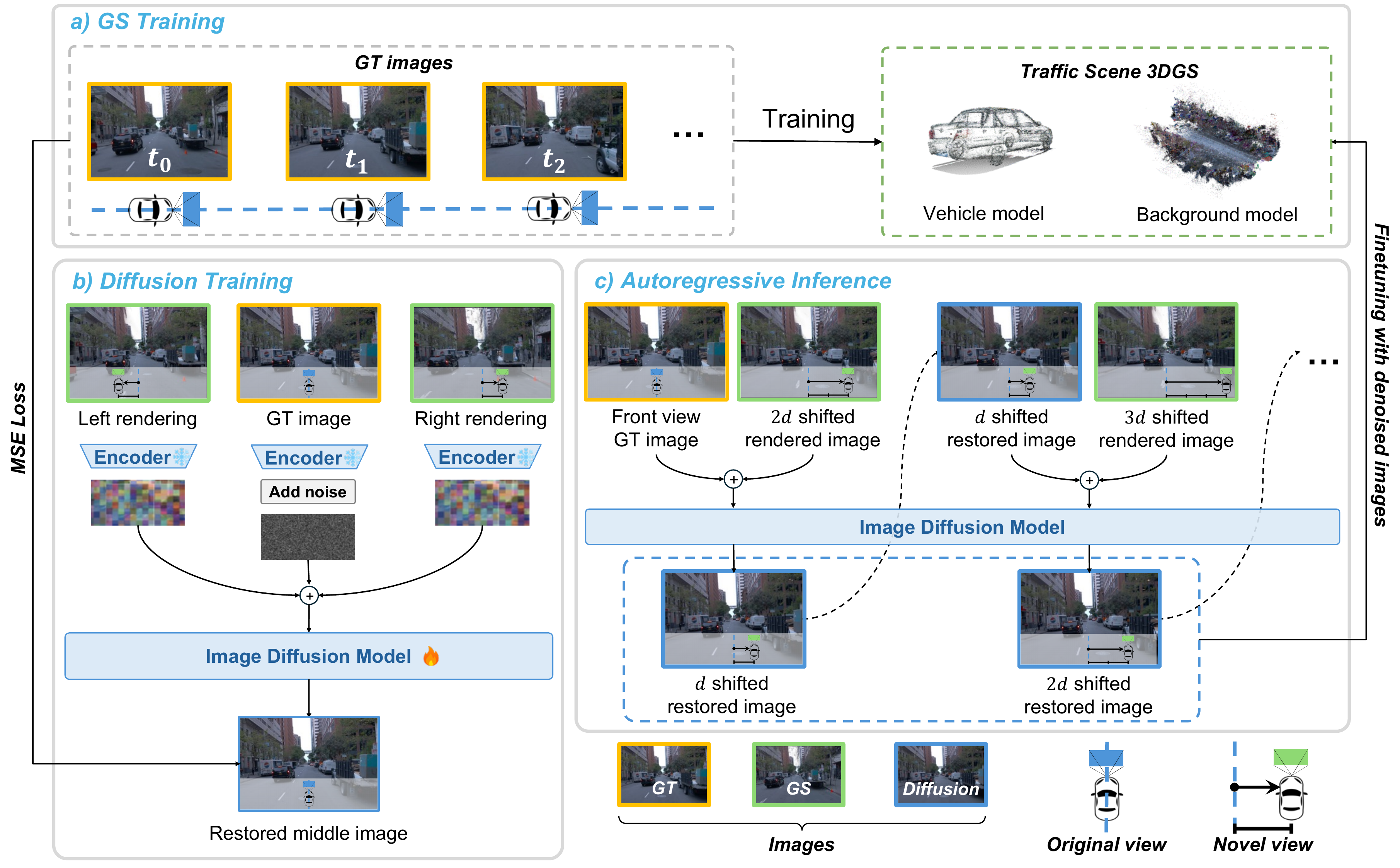}
    \caption{Novel-view restoration pipeline overview. a) The Gaussian Splatting (GS) model is trained separately for foreground vehicles and the background scene using ground truth (GT) images. b) Symmetric GS-rendered images are generated centered around the GT, and these symmetric data are used to train the diffusion model. c) Denoised novel view images are progressively generated via an autoregressive iterative process, and these images are then used to fine-tune the GS model.}
    \label{fig:pipeline test}
\end{figure*}

\paragraph{High-fidelity Closed-loop Simulation} High-fidelity closed-loop simulation frameworks integrate advanced visual rendering with sophisticated behavior control models for comprehensive interactive testing. For instance, DriveArena \cite{yang2024drivearena} employs controllable diffusion models for visual simulation, using lane lines and vehicle bounding boxes to constrain lane and vehicle positions. It incorporates LimSim \cite{wen2023limsim} for behavior control, fostering realistic traffic dynamics. Similarly, HugSim \cite{zhou2024hugsim} leverages 3D Gaussian Splatting for visual simulation, achieving controllable scene rendering by decoupling foreground vehicles and replacing them with pre-trained 3D car models to enhance visual fidelity and visibility. A rule-based behavior model handles action decision-making and trajectory planning, closing the simulation loop. While these pioneering efforts have significantly advanced high-fidelity closed-loop simulation, their visual realism is still constrained. In particular, 3DGS-based simulators often suffer from degraded novel-view rendering quality and struggle to maintain visual harmony during traffic editing. To mitigate these issues, recent diffusion-based models have been introduced to enhance novel-view rendering\cite{Ni2024ReconDreamerCW, yan2024streetcrafter, zhao2025recon} or perform realistic traffic editing\cite{podell2023sdxlimprovinglatentdiffusion, zhang2025scaling, ljungbergh2025r3d2realistic3dasset}. In this work, we explore the capability of a single diffusion model to support both tasks within a unified framework, aiming to jointly improve novel-view rendering quality and visual harmony during scene editing in closed-loop simulation.
\section{Methodology}
\label{sec:method}

Our work presents a framework for photorealistic traffic scene simulation combining 3D Gaussian Splatting (3DGS) \cite{kerbl3Dgaussians} for scene reconstruction with diffusion models for scene refinement. The methodology addresses two core challenges: (1) constructing a drivable 3D environment through differentiable Gaussian rendering that maintains visual fidelity across arbitrary viewpoints and trajectories; (2) modifying existing vehicles and inserting new vehicles while maintaining photorealism. 
In this section, we first introduce preliminaries for 3DGS and diffusion models (\cref{preliminaries}), then detail the training and inference pipeline of our diffusion model (\cref{training}–\cref{inference}), and finally present our approach for vehicle modeling and traffic simulation (\cref{insertion and simulation}).

\subsection{Preliminaries}
\label{preliminaries}

\textbf{3D Gaussian Splatting.}
3D Gaussian Splatting (3DGS) represents a scene using a set of anisotropic Gaussian primitives, enabling high-quality real-time rendering. Each Gaussian is parameterized by its 3D mean $\bm{\mu}$, scale $\bm{s}$, rotation $\bm{q}$, opacity $\alpha$, and view-dependent color encoded by spherical harmonics. The pixel color is obtained via alpha compositing of all overlapping Gaussians:
\begin{equation}
    C = \sum_{i\in \mathcal{N}} c_i \alpha_i \prod_{j=1}^{i-1}(1 - \alpha_j),
\end{equation}
where $\mathcal{N}$ denotes the set of Gaussians projected onto the pixel. The opacity $\alpha_i$ is computed from the projected 2D covariance
\begin{equation}
    \bm{\Sigma}' = \bm{J}\bm{W}\bm{\Sigma}\bm{W}^T \bm{J}^T,
\end{equation}
with $\bm{J}$ being the projection Jacobian and $\bm{W}$ the viewing transformation.

StreetGS~\cite{yan2024street} extends 3DGS to dynamic traffic scenes by modeling vehicles in local coordinates and transforming them into the world frame using time-dependent rigid motions:
\begin{align}
    \label{eqn:vehicle trans}
    \bm{\mu}_w(t) &= \bm{R}(t)\bm{\mu}_l + \bm{T}(t), \\
    \label{eqn:vehicle rots}
    \bm{R}_w(t) &= \bm{R}(t)\bm{R}_l,
\end{align}
where $\bm{R}(t)$ and $\bm{T}(t)$ denote the vehicle pose at time $t$. In this work, we build upon StreetGS models pretrained on ground-truth-view images.

\textbf{Diffusion Models.}
Diffusion models achieve strong performance in image generation and restoration by learning to reverse a gradual noising process. For traffic scenes, they are particularly effective at recovering missing details in novel-view rendering. We adopt Flux.1-dev~\cite{flux2024}, a state-of-the-art flow-matching diffusion model. Given a clean image $\bm{x}_0$, the forward process is defined as
\begin{align}
    \label{eqn:add noise}
    \bm{z}_t &= (1-t)\bm{x}_0 + t\bm{\epsilon}, \\
    \mathcal{L} &= \mathbb{E}_{\bm{\epsilon}\sim\mathcal{N}(0,\bm{I})}
    \|\bm{v}_\theta(\bm{z}_t, t) - (\bm{\epsilon} - \bm{x}_0)\|^2,
\end{align}
where $\bm{v}_\theta$ predicts the flow field at time $t$.

\subsection{Diffusion training}
\label{training}

\textbf{Data preparation}. Existing approaches employ various methods to simulate degraded renderings for training data pairs, including random mask augmentation \cite{Ni2024ReconDreamerCW} and applying Gaussian perturbations \cite{fan2024freesimfreeviewpointcamerasimulation}. While these approaches demonstrate reasonable performance, they present fundamental limitations: either the simulated degradation patterns fail to accurately capture real lateral view characteristics, or the data generation process involves complex artificial constructions.

Our approach leverages a pretrained 3DGS model $\mathcal{G}$ to generate training samples. Given a ground-truth image $I_0$ captured at camera pose $C_0$, we construct training inputs by laterally shifting the camera to symmetric positions $C_d$ and $C_{-d}$, where $d$ denotes the lateral displacement. The corresponding rendered views are obtained as
\begin{equation}
    I_d = \mathcal{G}(C_d), \quad I_{-d} = \mathcal{G}(C_{-d}).
\end{equation}
Each training sample therefore consists of an input pair $(I_d, I_{-d})$ and the central ground-truth target $I_0$, which naturally reflects the rendering degradation caused by lateral viewpoint shifts.

%forming a more accurate representation of the actual view degradation process compared to previous simulation methods.

\textbf{Training}. Our diffusion training process is illustrated in the left part of~\cref{fig:pipeline test}. For each training sample $(I_d, I_{-d}, I_0)$, where $I_d$ and $I_{-d}$ serve as the condition images and $I_0$ is the target image, we first encode the images into latent spaces represented as $\bm{z}_d, \bm{z}_{-d}, \bm{z}_0$ with a VAE encoder. Following Eq.~(\ref{eqn:add noise}), we obtain the noisy latent $\bm{z}_{0,t}$ by adding noise $\bm{\epsilon} \sim \mathcal{N}(0,\bm{I})$ to $\bm{z}_0$. The diffusion model $v_\theta$ then processes the concatenated latent representations $\bm{z}_d, \bm{z}_0, \bm{z}_{-d}$ with the training objective:
\begin{equation}
    \mathcal{L} = \mathbb{E}_{\bm{z}_0, \bm{\epsilon}, t} \left[\|\bm{v}_\theta([\bm{z}_{-d}; \bm{z}_{0,t}; \bm{z}_d], t) - (\bm{\epsilon} - \bm{z}_0)\|_2^2\right].
\end{equation}

The bilateral input structure enables the model to achieve superior reconstruction quality through multi-view consensus and complementary information fusion. By simultaneously processing both $I_d$ and $I_{-d}$, the model can: (1) identify and verify consistent features across views to establish robust geometric constraints, and (2) selectively combine the most reliable visual cues from each perspective when synthesizing missing regions. This architecture effectively addresses inherent ambiguities in single-view restoration, as it learns an adaptive fusion strategy that automatically weights view-specific evidence based on reliability, achieving significant improvements in both geometric consistency and texture fidelity.

%This approach effectively resolves ambiguities present in single-view reconstruction - areas that appear degraded or occluded in one view often contain valuable information in the other view. The model learns to automatically determine when to trust one view's evidence versus when to seek consensus, resulting in reconstructions that maintain better geometric accuracy while recovering more faithful texture details than single-view alternatives. Particularly for lateral view synthesis, this bilateral reference mechanism proves crucial as it preserves view-consistent structures while reducing artifacts that typically arise from viewpoint extrapolation.

\subsection{Diffusion inference and 3DGS refinement}
\label{inference}

\textbf{Diffusion inference}. The inference process follows an autoregressive view propagation scheme that progressively synthesizes novel views from known or restored neighbor views. Beginning with the ground truth image \( I_0 \), we first synthesize the rendered adjacent view \( \hat{I}_d \) through a diffusion process: starting from noise \( \epsilon \sim \mathcal{N}(0,1) \), we iteratively denoise \( z_{d,t} \) using the diffusion model \( \bm{v}_\theta([\bm{z}_0, \bm{z}_{d,t}, \bm{z}_{2d}], t) \), where the model leverages both the ground truth view \( \bm{z}_0 \) and the rendered view \( \bm{z}_{2d} \) to reconstruct the intermediate view \( \bm{z}_d \).
 Finally, we apply the VAE decoder to obtain the restored view \( \hat{I}_d \). This process then propagates outward in a chained manner - using the newly synthesized \( \hat{I}_d \) as input to generate \( \hat{I}_{2d} \) from the pair \( (\hat{I}_d, I_{3d}) \), and subsequently \( \hat{I}_{3d} \) from \( (\hat{I}_{2d}, I_{4d}) \), forming a robust autoregressive view propagation chain.

This iterative refinement benefits significantly from the ground truth initialization, where the high-quality \( I_0 \) serves as both anchor and information source throughout the propagation chain. The strong structural priors from \( I_0 \) not only guide initial view synthesis but continue to propagate through the sequence - each generated view inherits and refines these priors while serving as an improved starting point for subsequent neighbors, thereby maintaining robust geometric consistency across all synthesized views. This design enables the model to effectively disambiguate plausible content by leveraging both the propagated information from the ground truth and multiple consistent hypotheses from bidirectional context, ultimately achieving superior occlusion recovery and view consistency compared to single-pass generation approaches.

\textbf{Position-Accurate initialization}. While direct inference from bilateral views can generate intermediate images, this approach sometimes yields imprecise positional alignment in the synthesized middle view. Such positional inaccuracies may produce geometric inconsistencies when refining the 3D reconstruction model. To address this, we propose a noise initialization strategy where the diffusion process begins from denoising step $N_{\text{start}}$ instead of pure noise $\bm{\epsilon} \sim \mathcal{N}(0, \bm{I})$:
\begin{equation}
\label{eqn:noise init}
\bm{z}_{N_{\text{start}}} = (1-{\sigma_{N_{\text{start}}}})\bm{z}_0+{\sigma_{N_{\text{start}}}}\bm{\epsilon},
\end{equation}
where the noise scale $\sigma_{N_{\text{start}}}$ is sufficiently large. This initialization maintains two crucial properties: (1) strict preservation of the source image's global structural coherence, and (2) flexible synthesis of high-quality content in occluded regions. This balanced initialization leverages the encoded source geometry while allowing the diffusion model to hallucinate plausible details where visual evidence is absent, achieving both positional accuracy and visual realism in the synthesized views.

\textbf{3DGS Refinement}. After obtaining the restored novel view images through our diffusion-based synthesis pipeline, we employ these samples to refine the 3D Gaussian Splatting (3DGS) reconstruction model initially pretrained on ground-truth views. Following established practices in \cite{zhao2024drive, Ni2024ReconDreamerCW}, we optimize the model using a composite loss function with different components for ground-truth and novel views:
\begin{align}
\mathcal{L}_{\text{total}} &= \mathcal{L}_{\text{gt}} + \mathcal{L}_{\text{novel}}, \\
\mathcal{L}_{\text{gt}} &= \mathcal{L}_{\text{rgb}} + \lambda_1\mathcal{L}_{\text{ssim}} + \lambda_2\mathcal{L}_{\text{depth}}, \\
\mathcal{L}_{\text{novel}} &= \mathcal{L}_{\text{rgb}} + \lambda_1\mathcal{L}_{\text{ssim}},
\end{align}
where $\mathcal{L}_{\text{rgb}}$ measures the pixel-wise RGB reconstruction error, $\mathcal{L}_{\text{ssim}}$ enforces structural similarity preservation, and $\mathcal{L}_{\text{depth}}$ provides depth supervision.

\subsection{Vehicle insertion and simulation}
\label{insertion and simulation}

%\textbf{Vehicle insertion}. (one sentence like, for simulation, should create traffic, like place the vehicle as you want ...) some traffic reconstruction model \cite{yan2024street, chen2025omnire} separately model vehicles, enable translate and copy to create traffic. but, vehicle unseen view, hard to recover, as shown in fig xxx. HugSim \cite{zhou2024hugsim} use 3DRealCar \cite{du20243drealcar} to insert, possible solution, but unnatural light, texture.

%we follow, insert 3drealcar, but further use diffusion model to make it natural. (some inline equations, such as insert, obtain a image, diffusion model restore, and model fine-tune) ........ (finally conclude that we get a realistic, natural, complete vehicle model, foundation to simulate)

\textbf{Vehicle insertion}. For realistic traffic simulation, we require the ability to arbitrarily place and manipulate vehicles in the scene. Existing traffic reconstruction models \cite{yan2024street, chen2025omnire} separately model vehicles, enabling vehicle duplication and translation to create traffic flows. However, these models produce incomplete vehicle representations, particularly failing to reconstruct plausible geometry and textures for unseen views. While complete 3DRealCar \cite{du20243drealcar} assets used in HugSim \cite{zhou2024hugsim} provide full geometric coverage, they often exhibit unrealistic lighting and texture discontinuities with the surrounding scene.

% Our approach builds upon this foundation while addressing its limitations. Given an inserted 3DRealCar model $\mathcal{G}_v$, we place it at multiple positions and orientations to ensure comprehensive multiview restoration. For each configuration, we render the scene to obtain $I_{\text{insert}}^i$ and apply our diffusion model $\bm{v}_\theta([\bm{z}_{\text{insert}}, \bm{z}_t, \bm{z}_{\text{insert}}], t)$ to generate harmonized vehicle appearances $\tilde{I}_{\text{insert}}^i$, which exhibit improved lighting consistency, realistic material textures, and natural color blending with the surrounding scene. The diffusion model ensures that the vehicle's appearance adapts to environmental conditions, such as shadows and reflections, while preserving key geometric features. We then fine-tune $\mathcal{G}_v$ using these images with a composite $\mathcal{L}_{\text{rgb}} + \lambda\mathcal{L}_{\text{ssim}}$ loss, where optimization modifies only the vehicle's color attributes $\bm{c}_v$ and opacity $\alpha_v$ to preserve geometric integrity. The resulting refined model $\tilde{\mathcal{G}}_v$ maintains complete 3D structure while achieving photorealistic integration with the traffic scene, thus enabling flexible placement at arbitrary locations within the environment for high-fidelity simulation.

Our approach builds upon this foundation while addressing its limitations. Given an inserted 3DRealCar model $\mathcal{G}_v$, we place it at multiple positions and orientations within the scene to obtain diverse viewpoints. 
For each configuration, we render the scene to produce an initial inserted image $I_{\text{insert}}^i$. To harmonize the inserted vehicle with the background, we formulate traffic editing as an inpainting problem and directly reuse the same diffusion model without additional training. 
Specifically, we apply the diffusion model $\bm{v}_\theta([\bm{z}_{\text{insert}}, \bm{z}_t, \bm{z}_{\text{insert}}], t)$ within a RePaint-based framework~\cite{lugmayr2022repaintinpaintingusingdenoising}. 
At each denoising step, a binary mask corresponding to the inserted vehicle is applied: the latent features of the background region are reset to those of the original rendered image, while only the masked vehicle region is allowed to be updated. This iterative masking-and-denoising strategy enforces strict background consistency while enabling the vehicle appearance to progressively adapt to the surrounding environment. As a result, the diffusion model produces harmonized images $\tilde{I}_{\text{insert}}^i$ while preserving the original geometric structure of the inserted vehicle. 

We then fine-tune $\mathcal{G}_v$ using these images with a composite $\mathcal{L}_{\text{rgb}} + \lambda\mathcal{L}_{\text{ssim}}$ loss, where optimization modifies only the vehicle's color attributes $\bm{c}_v$ and opacity $\alpha_v$ to preserve geometric integrity. The resulting refined model $\tilde{\mathcal{G}}_v$ maintains complete 3D structure while achieving photorealistic integration with the traffic scene, thus enabling flexible placement at arbitrary locations within the environment for high-fidelity simulation.

\textbf{Traffic simulation}. 
With our photorealistic vehicle models $\tilde{\mathcal{G}}_v$ together with original scene vehicles, traffic behaviors can be simulated by directly controlling the motion parameters $\bm{R}(t)$ and $\bm{T}(t)$ defined in~\cref{eqn:vehicle trans,eqn:vehicle rots}. 
Our rendering framework is compatible with existing traffic control modules, including rule-based simulators~\cite{SUMO2018, Dosovitskiy17, Treiber_2000, wen2023limsim} for structured traffic flow, as well as learning-based trajectory generators~\cite{zhong2023languageguided, liu2025controllabletrafficsimulationllmguided, rowe2024ctrlsim, 10161463, jiang2024scenediffuser, tan2023language, feng2023trafficgen} for more complex and interactive behaviors. 
This compatibility enables closed-loop traffic simulation where diverse vehicle trajectories can be rendered with high visual fidelity, highlighting the potential of combining advanced traffic control with our photorealistic rendering.

% \textbf{Traffic simulation}. With our photorealistic vehicle models $\tilde{\mathcal{G}}_v$ and original scene vehicles, we precisely control the motion parameters $\bm{R}(t)$ and $\bm{T}(t)$ defined in~\cref{eqn:vehicle trans} and~\cref{eqn:vehicle rots} to simulate diverse traffic behaviors. For normal traffic flow control, we can integrate rule-based traffic simulators \cite{SUMO2018, Dosovitskiy17, Treiber_2000, wen2023limsim} to establish a real-time closed-loop simulation system where vehicle behaviors influence and respond to the evolving traffic conditions. This provides a stable and efficient simulation of routine traffic scenarios following standard driving rules. Besides, we also attempt to simulate the complex interactions in traffic scenes. We can utilize learning-based trajectory simulators \cite{zhong2023languageguided, liu2024controllabletrafficsimulationllmguided, rowe2024ctrlsim, 10161463, jiang2024scenediffuser, tan2023language, feng2023trafficgen} to generate realistic and complex vehicle trajectories and control the vehicle behaviors. With the powerful trajectory simulators and our high-fidelity rendering capabilities, we achieve highly realistic traffic simulations capable of representing diverse traffic scenes.
\section{Experiments}
\label{sec:exp}

In this section, we design and conduct a series of experiments to answer the following Critical Research Questions (RQs):

\begin{itemize}
    \item \textbf{RQ1:} Can the proposed symmetric auto-regressive mechanism effectively leverage Ground Truth (GT) guidance to refine novel view rendering? (\cref{exp:novel_view_rendering})
    \item \textbf{RQ2:} Does SymDrive, trained under a unified framework, achieve State-of-the-Art (SOTA) performance in the task of vehicle insertion? (\cref{exp:vehicle_insertion})
    \item \textbf{RQ3:} Is the realism of dynamic scenes rendered by SymDrive sufficient to support the decision-making processes of end-to-end autonomous driving agents?
\end{itemize}

% To validate our proposed framework, we conducted a series of comprehensive experiments. This section details the experimental setup, including the datasets, state-of-the-art baseline methods for comparison, and the evaluation metrics used to quantify performance. We then present results for novel view synthesis, enhanced vehicle insertion, and traffic simulation, supplemented by ablation studies analyzing the contributions of key components in our approach.

\begin{table}[t]
    \centering
    \caption{Performance Comparison on lateral shift 3m renderings. We \textbf{bold} the best result and \underline{underline} the second result.}
    \vspace{5pt}
    \renewcommand{\arraystretch}{1.3}
    \resizebox{0.48\textwidth}{!}{%
    \begin{tabular}{lccccc}
    \hline
    \textbf{Method} & \textbf{Extra condition} & \textbf{NTA-IoU $\uparrow$} & \textbf{NTL-IoU $\uparrow$} & \textbf{FID $\downarrow$} \\ \hline
    Street Gaussians \cite{yan2024street} & - & 0.498 & 53.19 & 130.75 \\
    FreeVS \cite{wang2024freevs} & - & 0.505 & 53.26 & 104.23 \\
    DriveDreamer4D \cite{zhao2024drive} & bbox\&map & 0.457 & 53.30 & 113.45 \\ 
    ReconDreamer \cite{Ni2024ReconDreamerCW} & bbox\&map & 0.539 & 54.58 & 93.56 \\ 
    ReconDreamer++$^{*}$ \cite{zhao2025recon} & bbox\&map & 0.566 & 56.89 & 75.22 \\
    Difix3D+ \cite{wu2025difix3d+} & - & \underline{0.578} & 56.94 & 84.12 \\
    ReconDreamer++$^{\dagger}$ \cite{zhao2025recon} & bbox\&map & 0.572 & \underline{57.06} & \textbf{72.02} \\
    \textbf{Ours} & - & \textbf{0.582} & \textbf{57.91} & \underline{74.82} \\
    \hline 
    \end{tabular}%
    }
    \par % 另起一段
    \vspace{0.3em}
    \raggedright 
    \scriptsize % 使用稍小的字号
    \hspace{1em} ReconDreamer++$^{*}$ denotes the standard reconstruction–generation pipeline. ReconDreamer++$^{\dagger}$ employs an auxiliary network to model geometrical modifications between ground-truth and novel views, which may introduce slight inconsistencies across viewpoints. In contrast, our approach aims to construct a consistent 3D representation across all viewing angles.

    \label{tab:results}
%\vspace{-15pt}
\vspace{-1em}
\end{table}

%ReconDreamer++(VD) employs an auxiliary network to learn residual corrections for Gaussian parameters between ground-truth and novel views. While this design choice helps achieve better metric alignment with diffusion outputs for novel views, it might potentially introduce slight inconsistencies across viewpoints. In contrast, our approach aims to build a consistent 3D representation across all viewing angles.

\begin{figure}[t]
    \centering
    \includegraphics[width=0.49\textwidth]{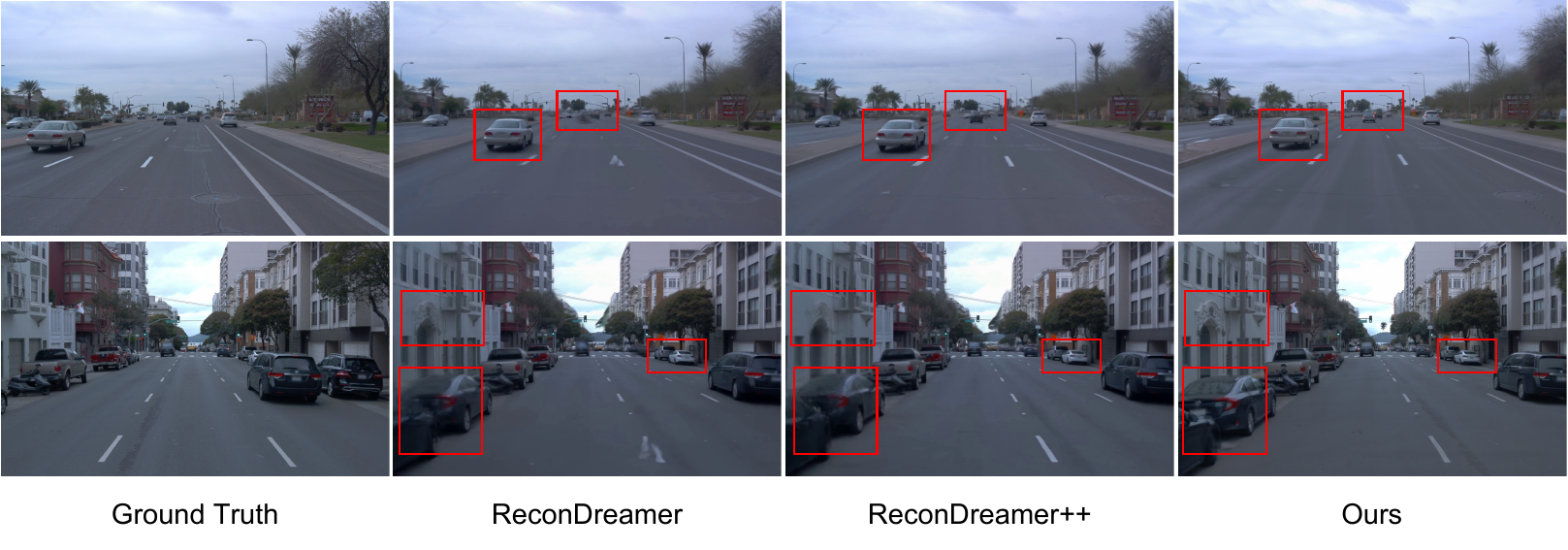}
    \caption{Qualitative comparison with ReconDreamer \cite{Ni2024ReconDreamerCW} and ReconDreamer++ \cite{zhao2025recon}.}
    % \caption{Qualitative comparison with ReconDreamer \cite{Ni2024ReconDreamerCW} and ReconDreamer++ \cite{zhao2025recon}. our approach better preserves fine near-field details and achieves more accurate reconstruction of road surface markings}
    \label{fig:recon_vis}
    \vspace{-1em}
\end{figure}

% \footnotetext{ReconDreamer++(VD) employs an auxiliary network to learn residual corrections for Gaussian parameters between ground-truth and novel views. While this design choice helps achieve better metric alignment with diffusion outputs for novel views, it might potentially introduce slight inconsistencies across viewpoints. In contrast, our approach aims to build a consistent 3D representation across all viewing angles.}
%between 3D Gaussian Splatting (3DGS) renderings from novel views and their denoised counterparts, aiming to improve data fitting and performance metrics. However, the denoised images may lose details from the Ground Truth (GT) images, compromising visual fidelity. Our method, by directly leveraging GT image information, generates textures with enhanced realism and richer detail.}

\subsection{Experiment Setup}
\label{sec:exp_setup}

\textbf{Dataset}. We conduct our experiments on the Waymo Open Dataset~\cite{Sun_2020_CVPR}, a large-scale autonomous driving dataset that provides high-quality, diverse sensor data captured in various urban environments. For quantitative evaluation of novel view synthesis, we follow~\cite{zhao2025recon} and select eight representative scenes, each containing 40 consecutive frames. As the NuScenes dataset\cite{nuscenes2019} lacks a unified and widely adopted benchmark for reconstruction-based diffusion novel view synthesis, we include qualitative visualizations on NuScenes in the supplementary material for completeness.

\textbf{Baselines.} For the novel view rendering task, we compare our method against representative state-of-the-art approaches, including the reconstruction model \textit{Street Gaussians}~\cite{yan2024street}, the generative model \textit{FreeVS}~\cite{wang2024freevs}, as well as hybrid methods that integrate generative restoration with 3D reconstruction, such as \textit{DriveDreamer4D}\cite{zhao2024drive}, \textit{ReconDreamer}\cite{Ni2024ReconDreamerCW}, \textit{ReconDreamer++}\cite{zhao2025recon}, and \textit{Difix3D+}\cite{wu2025difix3d+}. We additionally include qualitative comparisons with another hybrid state-of-the-art method, \textit{StreetCrafter}\cite{yan2024streetcrafter}. For the vehicle insertion harmonization task, we adopt the pixel-space editing model \textit{CosXL-Edit}~\cite{podell2023sdxlimprovinglatentdiffusion} and the novel view restoration method \textit{Difix3D+}~\cite{wu2025difix3d+} as baselines. We note that \textit{R3D2}\cite{ljungbergh2025r3d2realistic3dasset}, while currently representing the strongest performance in this setting, is not publicly available at the time of writing and thus cannot be included in our experimental comparisons.

% \textbf{Metrics}. Following \cite{zhao2024drive}, we employ three complementary metrics for novel view rendering assessment: (1) \textbf{NTA-IoU}  to quantify foreground object reconstruction quality, (2) \textbf{NTL-IoU} to evaluate lane marking fidelity, and (3) \textbf{FID} (Fréchet Inception Distance) \cite{NIPS2017_8a1d6947} to assess overall image quality and realism.

\textbf{Metrics}. Following~\cite{zhao2024drive}, we employ three complementary metrics for evaluating the novel view rendering task: \textbf{NTA-IoU} to quantify foreground object reconstruction quality, \textbf{NTL-IoU} to assess lane marking fidelity, and \textbf{FID} (Fréchet Inception Distance)~\cite{NIPS2017_8a1d6947} to measure overall image quality and realism. For the vehicle insertion harmonization task, we report \textbf{FID} scores computed before and after inserting 3DRealCar models\cite{du20243drealcar}, in order to evaluate the impact of insertion on image realism.

\textbf{Implementation details}. For the diffusion model component, we fine-tune the Flux.1-dev foundation model \cite{flux2024} using Low-Rank Adaptation (LoRA) \cite{hu2022lora} with rank 128. The model is trained on 4 NVIDIA A100 GPUs for 20,000 steps. During GT-guided autoregressive image restoration, we set the lateral shift distance $d$ to 0.5 meters per step. The denoising process runs for 50 steps, with noise initialization starting at $N_{\text{start}}=10$ in~\cref{eqn:noise init}. For the reconstruction model component, we adopt Street Gaussians \cite{yan2024street} and train it for 50,000 steps in total. Additionally, following\cite{zhao2025recon}, we integrate a ground model into the original framework, leveraging ground point cloud preprocessing for improved road surface reconstruction.

\begin{figure}[t]
    \centering
    \includegraphics[width=0.49\textwidth]{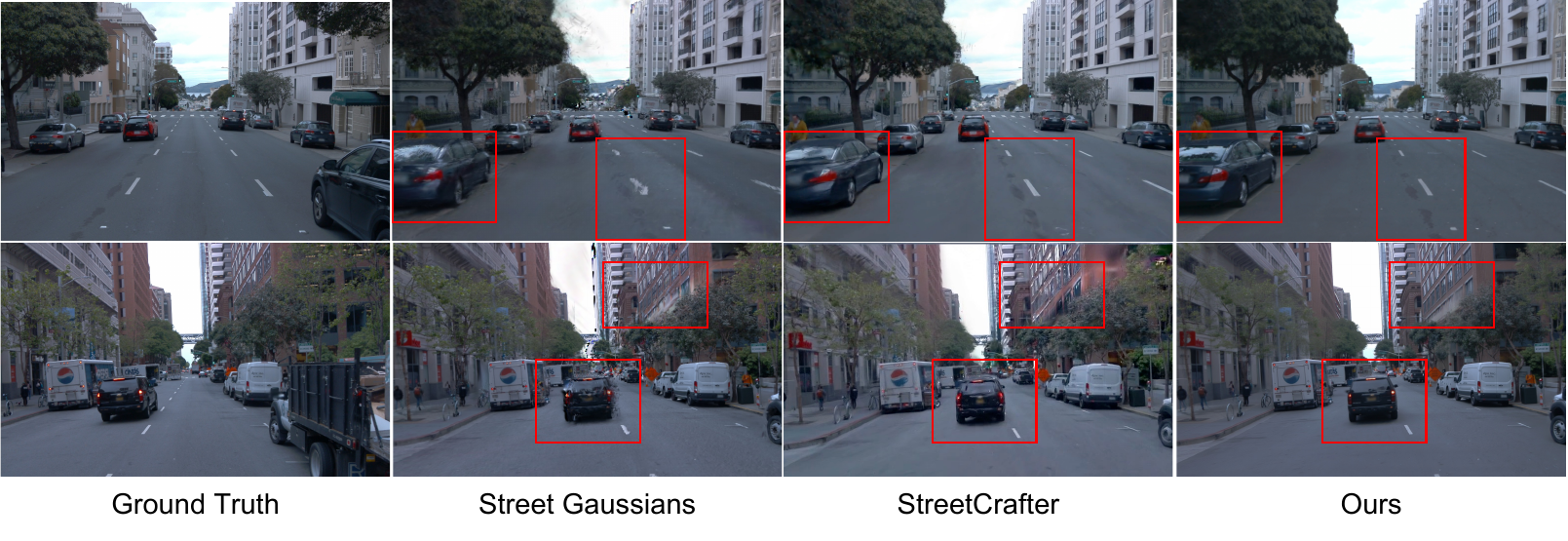}
    % \caption{Qualitative comparison with Street Gaussians \cite{yan2024street} and StreetCrafter \cite{yan2024streetcrafter}. our method shows significant improvements in both road surface representation (including lane markings and textures) and surrounding scene details such as traffic lights and roadside vehicles.}
    \caption{Qualitative comparison with Street Gaussians \cite{yan2024street} and StreetCrafter \cite{yan2024streetcrafter}. }
    \label{fig:sc_vis}
\end{figure}

\subsection{Novel view rendering}
\label{exp:novel_view_rendering}

% \textbf{Quantitative results}. Our method achieves competitive performance across all metrics. The base version (\textbf{Ours-base}) significantly outperforms existing hybrid methods, demonstrating superior reconstruction fidelity (higher NTA-IoU), more precise lane preservation (higher NTL-IoU), and better synthesized image quality (lower FID). Most importantly, compared to \textit{ReconDreamer}~\cite{Ni2024ReconDreamerCW}, which requires HD map data and agent bounding boxes as conditioning for single-view restoration, our framework achieves these improvements without requiring additional input information. This validates the effectiveness of our GT-guided design in achieving superior restoration quality.

% The ground-model enhanced version (\textbf{Ours-GM}) further improves these metrics, outperforming all PC-Hybrid pipelines while remaining competitive with VD-Hybrid approaches. Most notably, it achieves superior lane reconstruction accuracy (as reflected in NTL-IoU) without requiring explicit novel-view adaptation modules. 

\textbf{Quantitative results}. Our method achieves strong performance across all evaluation metrics, demonstrating superior reconstruction fidelity for both foreground objects and background structures, as well as high-quality synthesized images, compared to existing hybrid approaches. In particular, relative to \textit{ReconDreamer}\cite{Ni2024ReconDreamerCW} and ReconDreamer++\cite{zhao2025recon}, which relies on HD maps and agent bounding boxes as additional conditioning for single-view restoration, our framework consistently delivers improved results without requiring any extra input information. This highlights the effectiveness of our GT-guided design in enabling high-quality restoration.

\textbf{Qualitative results}. As shown in~\cref{fig:recon_vis} and~\cref{fig:sc_vis}, our method demonstrates superior novel view rendering quality. Compared to \textit{ReconDreamer}~\cite{Ni2024ReconDreamerCW} and \textit{ReconDreamer++}~\cite{zhao2025recon}, our approach better preserves fine near-field details and achieves more accurate reconstruction of road surface. The advantages are particularly evident when compared to \textit{StreetCrafter}~\cite{yan2024streetcrafter}, where our method shows significant improvements in both road surface representation (including lane markings and textures) and surrounding scene details such as traffic lights and roadside vehicles.

\begin{table}[!t]
\centering
\caption{Ablation study on noise initialization step $N_{start}$ and auto-regressive step size $d$. ($d=0.0$ m denotes direct single-view restoration for novel views)}
\vspace{5pt}
\renewcommand{\arraystretch}{1.3} % 增加行间距
\resizebox{0.49\textwidth}{!}{
\begin{tabular}{lcccc|ccc}
\hline
\multirow{2}{*}{} & \multicolumn{4}{c|}{\textbf{Start Denoising Step $N_{start}$}}                                               & \multicolumn{3}{c}{\textbf{Step Size $d$ (m)}}                                          \\ \cline{2-8}
& \textbf{0}             & \textbf{5}             & \textbf{10}            & \textbf{15}             & \textbf{0.0}             & \textbf{0.5}             & \textbf{1.0}             \\ \hline
\textbf{NTL-IoU}  $\uparrow$  & 56.76                  & 57.32                  & 57.91                  & 57.24                  & 56.83                  & 57.91                  & 57.04                  \\ 
\hline
\end{tabular}
}
\label{tab:ablation_study} % 添加标签以便引用
\end{table}

% \begin{table}[t]
%     \centering
%     \begin{minipage}[t]{0.48\textwidth}
%         \centering
%         \caption{Ablation study on noise initialization step $T_{\text{start}}$}
%         \vspace{5pt}
%         \setlength{\tabcolsep}{20pt}
%         \begin{tabular}{cc}
%             \hline
%             $T_{\text{start}}$ & NTL-IoU $\uparrow$ \\
%             \hline
%             0 & 55.76 \\
%             5 & 56.12 \\
%             10 & \textbf{56.38} \\
%             15 & 55.98 \\
%             \hline
%         \end{tabular}
%     \end{minipage}
%     \hfill
%     \begin{minipage}[t]{0.48\textwidth}
%         \centering
%         \caption{Ablation study on autoregressive step size $d$. $d=0$ means single-view restoration.}
%         \vspace{10pt}
%         \setlength{\tabcolsep}{20pt}
%         \begin{tabular}{cc}
%             \hline
%             Step Size & NTL-IoU $\uparrow$ \\
%             \hline
%             0.0m & 55.46 \\
%             0.5m & \textbf{56.38} \\
%             1.0m & 55.74 \\
%             \hline
%         \end{tabular}
%     \end{minipage}
% \end{table}

\textbf{Ablation Study}. We conduct ablation studies on two key components of our approach: the noise initialization strategy and the auto-regressive step size selection. Performance is measured using the NTL-IoU metric, which jointly assesses positional accuracy and rendering quality of the road surface, providing a comprehensive evaluation of our design choices. As shown in~\cref{tab:ablation_study}, the configuration without noise initialization ($N_{\text{start}}=0$) performs poorly due to the lack of initial guidance, leading to positional inaccuracies. While moderate noise initialization ($N_{\text{start}}=5,10$) improves performance, excessive initialization steps ($N_{\text{start}}=15$) degrade results, because it prevents effective correction of 3DGS rendering artifacts. Regarding the auto-regressive step size $d$, single-view rendering without GT guidance ($d=0.0$ m) yields suboptimal performance due to limited contextual information. A moderate step size of $d=0.5$m achieves the best balance, while larger steps ($d=1.0$ m) introduce artifacts from more distant, lower-quality renderings, resulting in slightly worse performance. These results underscore the effectiveness of our noise initialization strategy and GT-guided auto-regressive design.

\subsection{Vehicle insertion and simulation}
\label{exp:vehicle_insertion}

\begin{table}[t]
\centering
\renewcommand{\arraystretch}{1.2}
\caption{Comparison of vehicle insertion FID with different methods.}
\begin{tabular}{l l c}
\hline
\textbf{Method} & \textbf{Model Capability} & \textbf{FID $\downarrow$} \\
\hline
3DRealCar Insert & - & 41.27 \\
Difix3D & novel view restoration & 53.64 \\
CosXL-Edit & pixel to pixel image edit & 46.54 \\
Ours & - & \textbf{32.60} \\
\hline
\end{tabular}
\label{tab:insert_fid}
\end{table}

\textbf{Vehicle insertion}. As summarized in~\cref{tab:insert_fid}, our method achieves the best overall performance for vehicle insertion harmonization, indicating more realistic and coherent integration of inserted vehicles into complex driving scenes. Notably, although Difix3D+ is effective for novel view restoration, it is not designed for direct harmonization: during insertion, its diffusion-based restoration tends to simultaneously modify both foreground vehicles and background regions (e.g., brightening the background while darkening the inserted vehicles), leading to worse FID after harmonization. In contrast, our approach explicitly supports vehicle insertion within a unified framework. These results highlight the effectiveness of our design in supporting both novel view synthesis and vehicle harmonization within a single model.

\begin{figure}[t]
    \centering
    \includegraphics[width=0.49\textwidth]{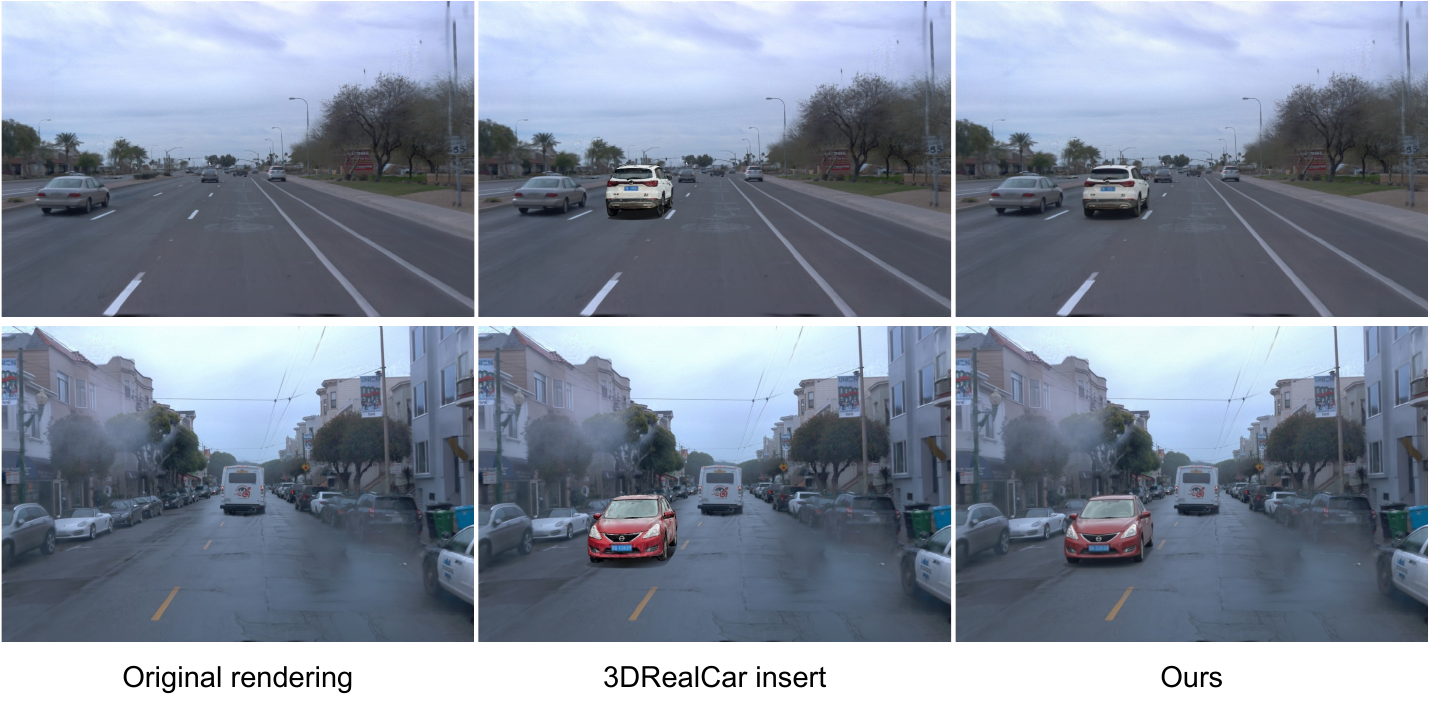}
    % \caption{Qualitative comparison with Street Gaussians \cite{yan2024street} and StreetCrafter \cite{yan2024streetcrafter}. our method shows significant improvements in both road surface representation (including lane markings and textures) and surrounding scene details such as traffic lights and roadside vehicles.}
    \caption{Qualitative results of vehicle insertion and harmonization.}
    \label{fig:insert_harm_vis}
\end{figure}

The visualization of vehicle insertion is shown in~\cref{fig:insert_harm_vis}. We demonstrate the flexible insertion of multiple vehicles into complex traffic scenes. Compared with HugSim~\cite{zhou2024hugsim}, which directly inserts pre-trained 3D vehicle models, our approach further enhances the results by first restoring the rendering through a diffusion model and then fine-tuning the vehicle appearance to precisely match the target scene. As shown in the figure, our method produces inserted vehicles with better background alignment, more detailed textures, and more natural lighting and color matching. The diffusion-based refinement process effectively bridges the domain gap between synthetic vehicle models and real-world scenes, preserving realistic interactions with environmental lighting while maintaining accurate perspective and scale.

% \textbf{Traffic Simulation.} As shown in~\cref{fig:simulation}, we demonstrate the capability of our simulation framework to model various traffic scenarios. In~\cref{fig:simulation} (a), we construct a denser traffic scene compared to the original environment and utilize SUMO~\cite{SUMO2018} for simulating dense but normal traffic flow. In addition, we also model complex driving behaviors. As shown in~\cref{fig:simulation} (b), we employ a learning-based controllable trajectory simulation method~\cite{zhong2023languageguided} to generate high-risk maneuvers such as aggressive lane-changing and overtaking behaviors. These challenging scenarios are particularly important for thoroughly evaluating autonomous driving systems. By seamlessly integrating our high-fidelity traffic scene editing and photorealistic rendering capabilities with various trajectory simulators, we achieve highly realistic and adaptable traffic simulation.

\begin{figure}[t]
    \centering
    \includegraphics[width=0.49\textwidth]{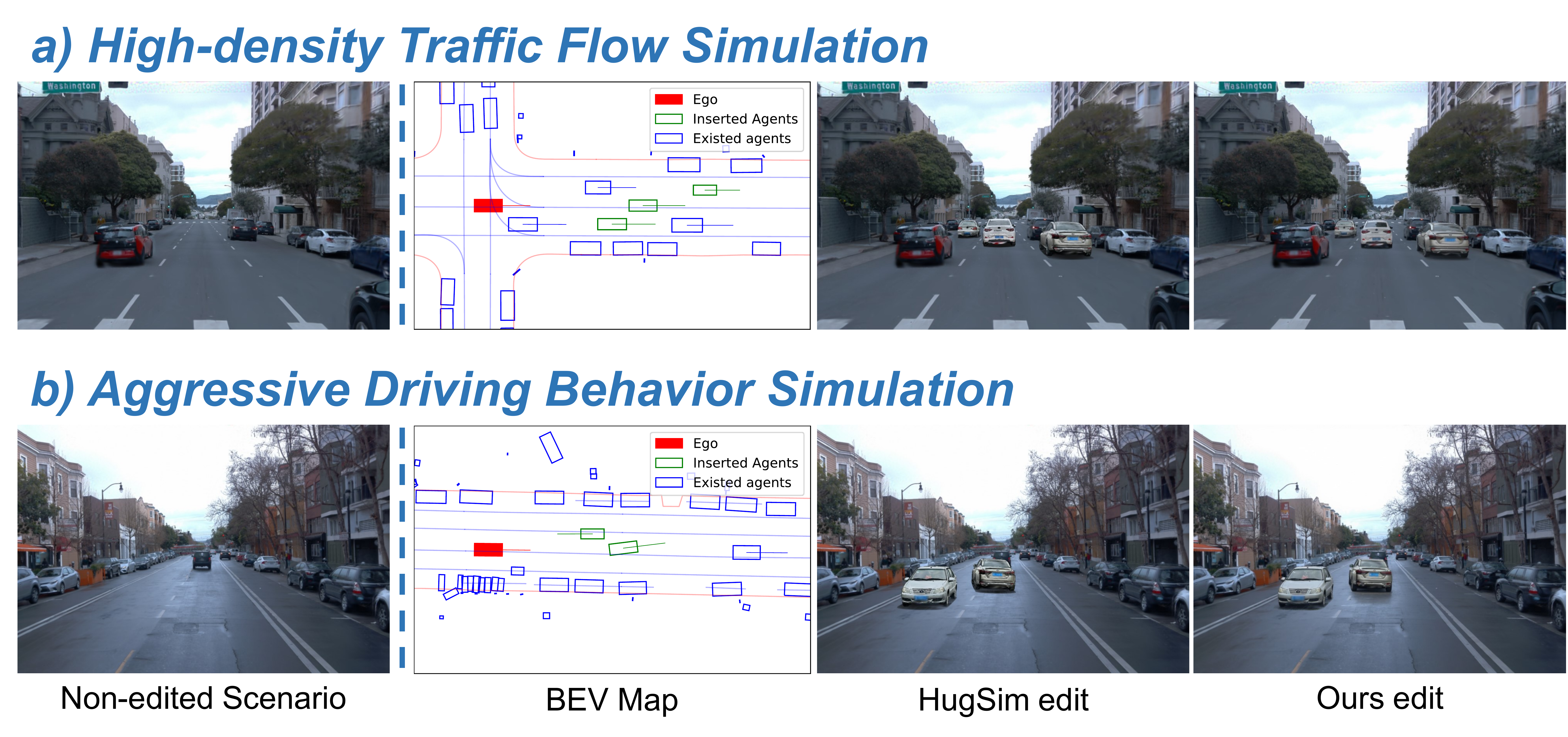}
    % \caption{Illustration of high-fidelity closed-loop simulation. a) Insert multiple pretrained 3D vehicles to create a high-density driving environment. b) Control background vehicles to perform aggressive maneuvers, creating corner case driving scenarios. Fine-tuned with our method, the 3D vehicle model integrates more harmoniously with the background environment in color and brightness compared to direct insertion of a pre-trained model~\cite{zhou2024hugsim}. Additionally, the vehicle glass reflections appear more realistic.}
    \caption{Illustration of high-fidelity closed-loop simulation}
    \label{fig:simulation}
\end{figure}

\begin{figure}[t]
    \centering
    \includegraphics[width=0.49\textwidth]{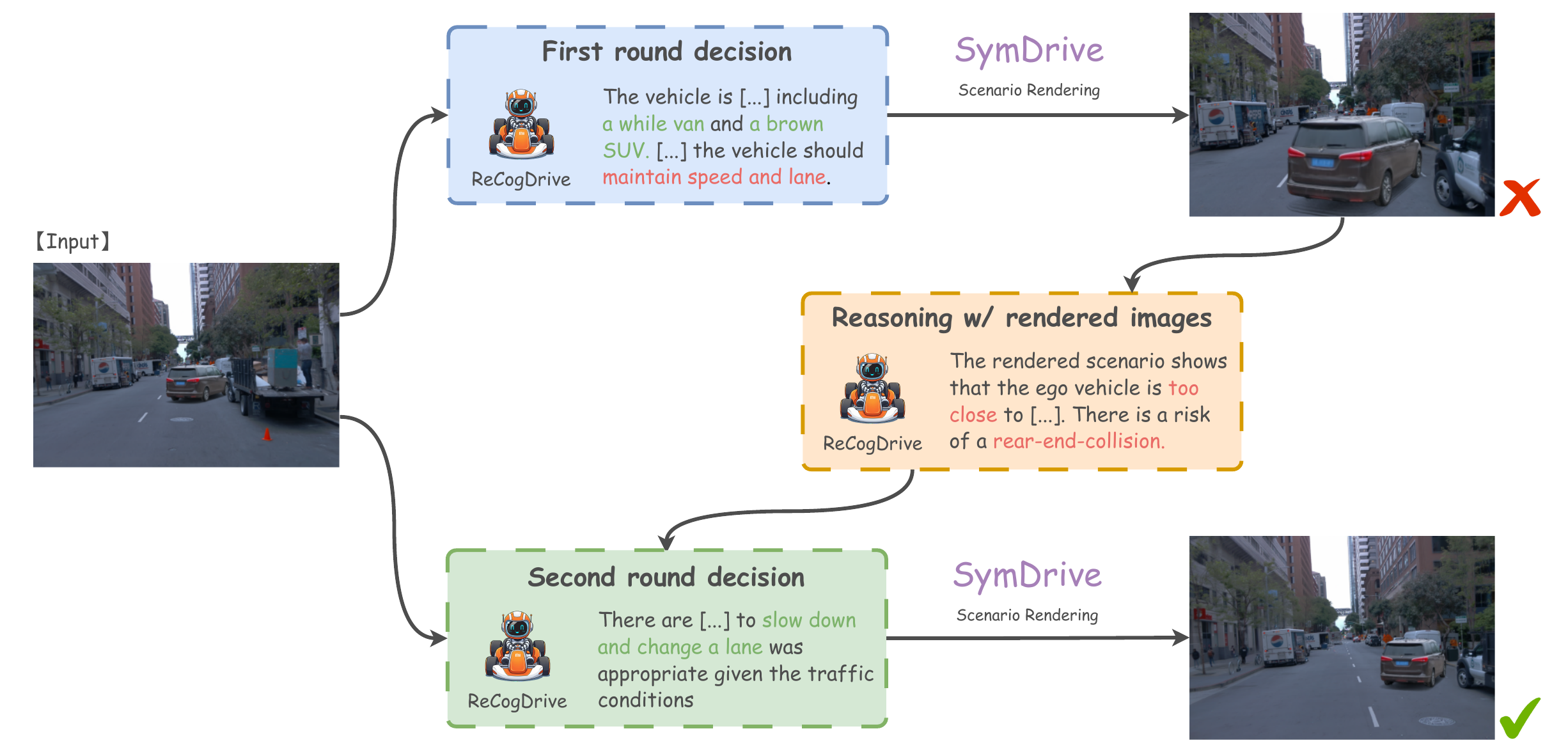}
    \caption{Example of VLM planning and reasoning within our simulation environment.}
    \label{fig:vlm_sim}
    \vspace{-1em}
\end{figure}

\textbf{Traffic Simulation.} As shown in~\cref{fig:simulation}, our simulation framework can model diverse traffic scenarios. In~\cref{fig:simulation}(a), we create denser traffic using SUMO~\cite{SUMO2018} to simulate normal flow, while in~\cref{fig:simulation}(b), we generate high-risk maneuvers such as aggressive lane changes and overtaking using a learning-based controllable trajectory simulator~\cite{zhong2023languageguided}. These challenging scenarios are crucial for evaluating autonomous driving systems. Additionally, to demonstrate the applicability of our framework in autonomy evaluation, we integrate our work with an end-to-end vision–language driving model\cite{li2025recogdrivereinforcedcognitiveframework}  in a closed-loop testing pipeline. As illustrated in Fig.~\ref{fig:vlm_sim}, the VLM first produces an initial driving decision based on the real input view. Leveraging our simulator’s ability to render photorealistic traffic from multiple viewpoints, we generate scenarios that reflect the potential consequences of the initial plan. We further attempt to feed simulated failure cases—such as scenes with unsafe proximity to preceding vehicles—back into the VLM to evaluate its reasoning capability, and find that the model is able to recognize risks and adjust its decisions appropriately. These results demonstrate the practical utility of our simulation framework for downstream autonomous driving tasks.

% we conduct experiments with a driving vision-language model\cite{li2025recogdrivereinforcedcognitiveframework} in our simulated environment: the VLM can successfully imagine intended behaviors and correctly recognize scene objects, demonstrating both the photorealism of our rendering and the practical utility of our simulation framework for downstream autonomous driving tasks.

We also study the rendering time of our framework under different traffic densities. As shown in Table~\ref{tab:fps_comparison}, our method maintains real-time performance under typical vehicle counts, demonstrating the efficiency and scalability of our simulation and rendering pipeline.

\begin{table}[t]
  \centering
  \caption{FPS Comparison under Different Settings}
  \label{tab:fps_comparison}
  \resizebox{0.49\textwidth}{!}{
  \renewcommand{\arraystretch}{1.05}
  \begin{tabular}{l c cccc}
    \toprule
    \multirow{2}{*}{Vehicle Type} & 
    \multirow{2}{*}{\#GS per Vehicle} & 
    \multicolumn{4}{c}{FPS at Different Vehicle Counts} \\
    \cmidrule(lr){3-6}
     &  & 5 & 10 & 25 & 50 \\
    \midrule
    Existed vehicle  & $\sim$4k    & 105.38 & 104.04 & 103.01 & 100.96 \\
    Inserted vehicle & $\sim$100k  & 96.09  & 80.84  & 54.38  & 37.30  \\
    \bottomrule
  \end{tabular}
  }
\end{table}

\vspace{-1.0mm}
% \vspace{-1.0mm}

\section{Conclusion}\label{sec:conclusion}

In this paper, we presented a novel framework for high-fidelity closed-loop autonomous driving simulation, addressing key challenges in novel view rendering and traffic controllability. Our approach features an auto-regressive novel view denoising algorithm that leverages ground truth images as priors, effectively enhancing rendering fidelity without requiring additional data. To ensure controllable background traffic, we decoupled foreground and background entities, integrating a traffic flow controller with high-quality 3DRealCar assets to enable dense and diverse traffic simulation.

Despite these advancements, limitations persist regarding distant objects, where sparse pixel representation hampers the efficacy of image-based priors, leading to temporal inconsistencies. Future work will explore video-diffusion models to improve long-range consistency and inference efficiency. Additionally, while our current system manages traffic patterns, it lacks rigid physics-based collision constraints. We plan to integrate a robust physics engine to further elevate the simulation's realism and safety validation capabilities.

\bibliographystyle{IEEEtran}
 % \vspace{-2.0mm}
%\bibliography{main}
%\bibliographystyle{unsrt}
\bibliography{main}

\end{document}